\documentclass[letterpaper, 10 pt, conference]{ieeeconf}  

\IEEEoverridecommandlockouts                              

\title{\LARGE \bf
Component-aware Unsupervised Logical Anomaly Generation for Industrial Anomaly Detection
}

\author{Xuan Tong$^{1}$, Yang Chang$^{1}$, Qing Zhao$^{1}$, Jiawen Yu$^{1}$, Boyang Wang$^{1}$, Junxiong Lin$^{1}$, Yuxuan Lin$^{1}$,\\ Xinji Mai$^{1}$, Haoran Wang$^{1}$, Zeng Tao$^{1}$, Yan Wang$^{1}$ and Wenqiang Zhang$^{2,*}$
\thanks{$^{1}$Xuan Tong, Yang Chang, Qing Zhao, Jiawen Yu, Boyang Wang, Junxiong Lin, Yuxuan Lin, Xinji Mai, Haoran Wang, Zeng Tao, Yan Wang are with Shanghai Engineering Research
Center of AI \& Robotics, Academy for Engineering \& Technology, Fudan University, Shanghai, China. {\tt\small\{xtong23, ychang24, jwyu23, bywang22, jxlin23, xjmai23, hrwang23\}@m.fudan.edu.cn, \{zhaoq19, ztao19, yanwang19\}@fudan.edu.cn}}%
\thanks{$^{2}$Wenqiang Zhang is with Engineering Research Center of AI \& Robotics, Ministry of Education, Academy for Engineering \& Technology, Fudan University, Shanghai, China.{\tt\small wqzhang@fudan.edu.cn}}%
\thanks{*Corresponding author: Wenqiang Zhang.}
}

\usepackage{amssymb}
\usepackage{amsmath}
\usepackage{array}
\usepackage{booktabs}
\usepackage[table,xcdraw]{xcolor}
\usepackage{tabularx}
\usepackage{adjustbox}
\usepackage{multirow}
\usepackage{cite}

\begin{document}

\maketitle
\thispagestyle{empty}
\pagestyle{empty}

\begin{abstract}
Anomaly detection is critical in industrial manufacturing for ensuring product quality and improving efficiency in automated processes. The scarcity of anomalous samples limits traditional detection methods, making anomaly generation essential for expanding the data repository. However, recent generative models often produce unrealistic anomalies increasing false positives, or require real-world anomaly samples for training. In this work, we treat anomaly generation as a compositional problem and propose ComGEN, a component-aware and unsupervised framework that addresses the gap in logical anomaly generation. Our method comprises a multi-component learning strategy to disentangle visual components, followed by subsequent generation editing procedures. Disentangled text-to-component pairs, revealing intrinsic logical constraints, conduct attention-guided residual mapping and model training with iteratively matched references across multiple scales. Experiments on the MVTecLOCO dataset confirm the efficacy of ComGEN, achieving the best AUROC score of 91.2\%. Additional experiments on the real-world scenario of Diesel Engine and widely-used MVTecAD dataset demonstrate significant performance improvements when integrating simulated anomalies generated by ComGEN into automated production workflows.

\end{abstract}


\section{INTRODUCTION}

Visual anomaly detection is fundamental to ensuring quality control in production and serves as a key enabler of industrial automation and evolution of intelligent manufacturing systems \cite{mei2018unsupervised,long2021fabric,yu2024multi}. Although existing methods \cite{zavrtanik2021draem,zhang2023destseg,deng2022anomaly,roth2022towards} rely on unsupervised learning, the lack of understanding of anomalies makes building a robust detection system difficult. Due to the scarcity of anomalous data, recent studies have shown that generating highly realistic anomalous samples can significantly improve the automation efficiency of downstream tasks such as anomaly detection \cite{zhang2021defect,duan2023few,hu2024anomalydiffusion,jiang2024cagen}.

\begin{figure}[t]
\centering
\includegraphics[width=1\columnwidth]{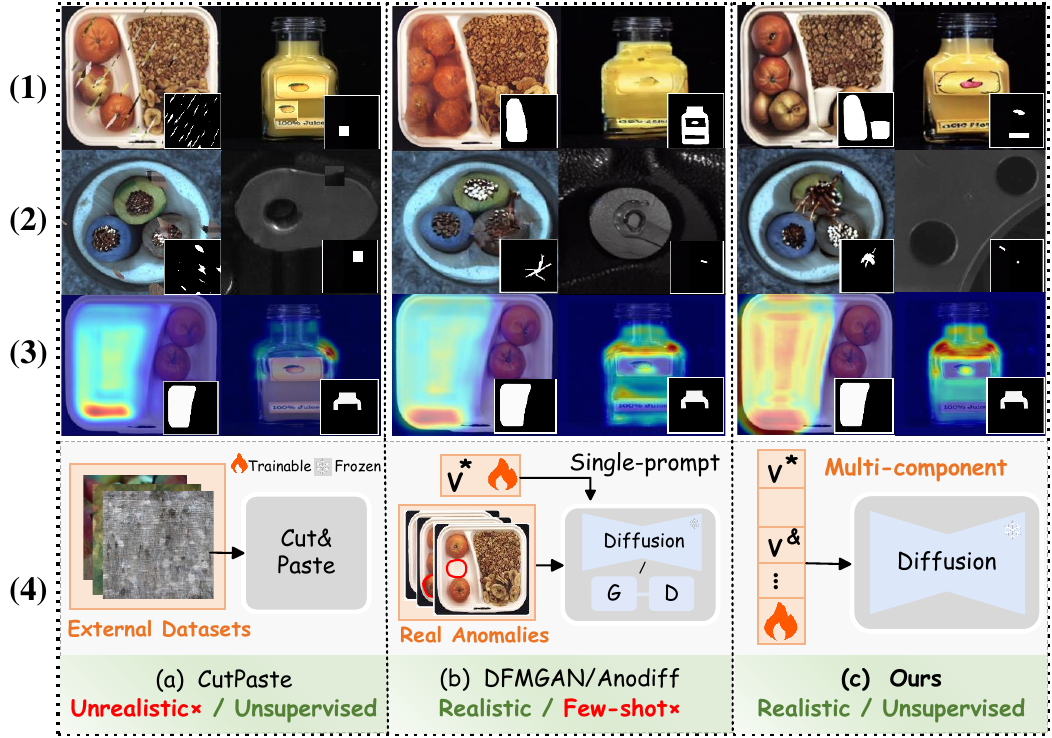} 
\caption{Comparison between existing anomaly generation methods and ours. \textbf{(1)} Anomaly generation results on MVTecLOCO dataset \cite{bergmann2022beyond}; \textbf{(2)} Anomaly generation results on MVTecAD dataset \cite{bergmann2019mvtec} and real-world Diesel Engine; \textbf{(3)} Anomaly localization results on MVTecLOCO (ground-truth masks in lower-right corner); \textbf{(4)} Comparison of network architectures.}
\label{fig1}
\end{figure}

To address this challenge, various studies have focused on visual anomaly generation
as illustrated in Fig. \ref{fig1}. Traditional methods \cite{zavrtanik2021draem,li2021cutpaste} produce unrealistic anomalies (Fig. \ref{fig1}a) by randomly cropping and pasting patterns from external datasets or the same image. Most recently, several few-shot generation models (Fig. \ref{fig1}b) have been trained using real anomalies to improve the realism of generated anomalies. First, GAN-based ways \cite{duan2023few} are prone to instability during training, often collapsing when faced with complex anomalous patterns. Second, Diffusion-based ways \cite{jiang2024cagen} utilize prior information of pretrained Latent diffusion models (LDMs) \cite{rombach2022high}. However, since there is a huge gap between industrial images and natural images \cite{lee2024text}, simply inputting natural text causes color or shape distortions during anomaly generation. Therefore, the most advanced method, AnoDiff \cite{hu2024anomalydiffusion}, jointly optimizes a single learnable prompt and a spatial encoder to achieve the generative goal. While it performs well with structural anomalies emphasizing local damages, Fig. \ref{fig1}b shows that when dealing with logical anomalies that violate underlying constraints, the generated anomalies do not align accurately with anomaly masks. To summarize, these generative models are either unsupervised but unrealistic, or they offer realistic generation but are biased towards structural anomalies generation and require real anomalous data.

In this work, we propose an unsupervised and component-aware anomaly generation framework ComGEN, which fills the gap in logical anomaly generation. Innovatively, given several normal industrial images, we assume that logical anomaly generation can be considered as a compositional problem, which involves changes in the layouts of multiple objects within an image (e.g., quantity and position). 
 
Based on this assumption, ComGEN employs a multi-component learning strategy to bridge text and components by disentangling image regions through multiple learnable embeddings, aligning each token's attention to its corresponding object region. Coupled with two generation editing approaches: prompt modifications and low-density sampling, further enhances generative quality. The generation process iterates with a memory-based reference association module to identify the closest normal neighbors to the anomalies. Between anomalous and normal images, their residual attention of each token maps to distortions in each component, producing interpretable anomaly masks. While their multi-scale differential features implicitly indicating anomalies, are aggregated into U-Net structure to facilitate cross-scale information exchange. Our key contributions are as follows:
\begin{itemize}
    \item A new perspective of treating logical anomaly generation as a compositional problem. We propose Multi-Component Learning (MCL) strategy to disentangle visual components in an unsupervised way, coupled with two generation editing approaches Prompt Modifications (PM) and Low-density Sampling (LS) to generate realistic and diverse anomalies beyond test set.
    \item A mask integrator combining iterative Anomaly-to-Reference Neighbor Association (RNA) and text-to-component Residual Mapping (RM) produces precise masks and implicit anomalous information, which guides Cross-Scale Difference Aggregation Module (CSDA) for accelerated anomaly detection learning.  
    \item Our framework ComGEN achieves advanced performance with interpretable results on MVTecLOCO \cite{bergmann2022beyond} dataset. Applications on the  real-world diesel engine industry and MVTecAD \cite{bergmann2019mvtec} dataset further demonstrate the model's scalability and potential in manufacturing. 
\end{itemize}

\section{RELATED WORKS}
Detecting approaches are divided into reconstruction-based, embedding-based and generation-based. Firstly, reconstruction-based methods \cite{zavrtanik2021draem,zavrtanik2021reconstruction} reconstruct normal images in the training stage, assuming that the model would reconstruct abnormal images with a large error, which is frequently contradicted during the test. Secondly, embedding-based methods \cite{ liu2023simplenet,zhang2023destseg,deng2022anomaly,roth2022towards} usually use a pre-trained network on ImageNet \cite{deng2009imagenet} to capture the high-level features of images. The anomaly score is then calculated by measuring the distance between the test sample and normal samples in the feature space. But industrial image features are different from natural images, so that directly using pre-trained features may cause a mismatch problem. Thirdly, generation-based methods \cite{niu2020defect,zhang2021defect,duan2023few,hu2024anomalydiffusion,jiang2024cagen} generate anomalous images to simulate potential deviations from the normal distribution as negative samples, which help the network learn to recognize and differentiate anomalous patterns more effectively.

Earliest generation-based methods DRAEM \cite{zavrtanik2021draem}, CutPaste \cite{li2021cutpaste} and NSA  \cite{schluter2022natural} augment normal samples by introducing abnormal patterns from patches within the same images or external texture dataset \cite{dtd2014accuracy}. But their generated samples often lack realism and diversity. Then GAN-based models like SDGAN \cite{niu2020defect}, DefectGAN \cite{zhang2021defect}, and DFMGAN \cite{duan2023few} train generative adversarial networks (GANs) \cite{goodfellow2020generative}  to produce anomalies by learning from real anomalies. Recently, text-guided approaches have emerged with the advancement of LDMs \cite{rombach2022high}. Anodiff \cite{hu2024anomalydiffusion} disentangles spatial information from anomaly appearance. CAGEN \cite{jiang2024cagen} combines the features of real anomalous features with normal samples using ControlNet \cite{zhang2023adding}.  Different with our method, these few-shot methods require real anomalies, and pay more attention on structural anomalies while ignore the logical anomalies.

Existing unsupervised generation-based methods towards logical anomalies include LogicalAL \cite{zhao2024logical} and GRAD \cite{dai2024generating}. LogicalAL manipulates edges and converts the modified edge map into image by using edge-to-image generator. GRAD removes self-attention and reduces network depth of U-Net architecture in the diffusion process to obtain local anomaly patterns. However, their generated anomalies visually differ from real anomalies and lack semantic interpretation, failing to capture the true nature of abnormal patterns which influence the performance of downstream detection task.

\begin{figure*}[t]
\centering
\includegraphics[width=1\textwidth]{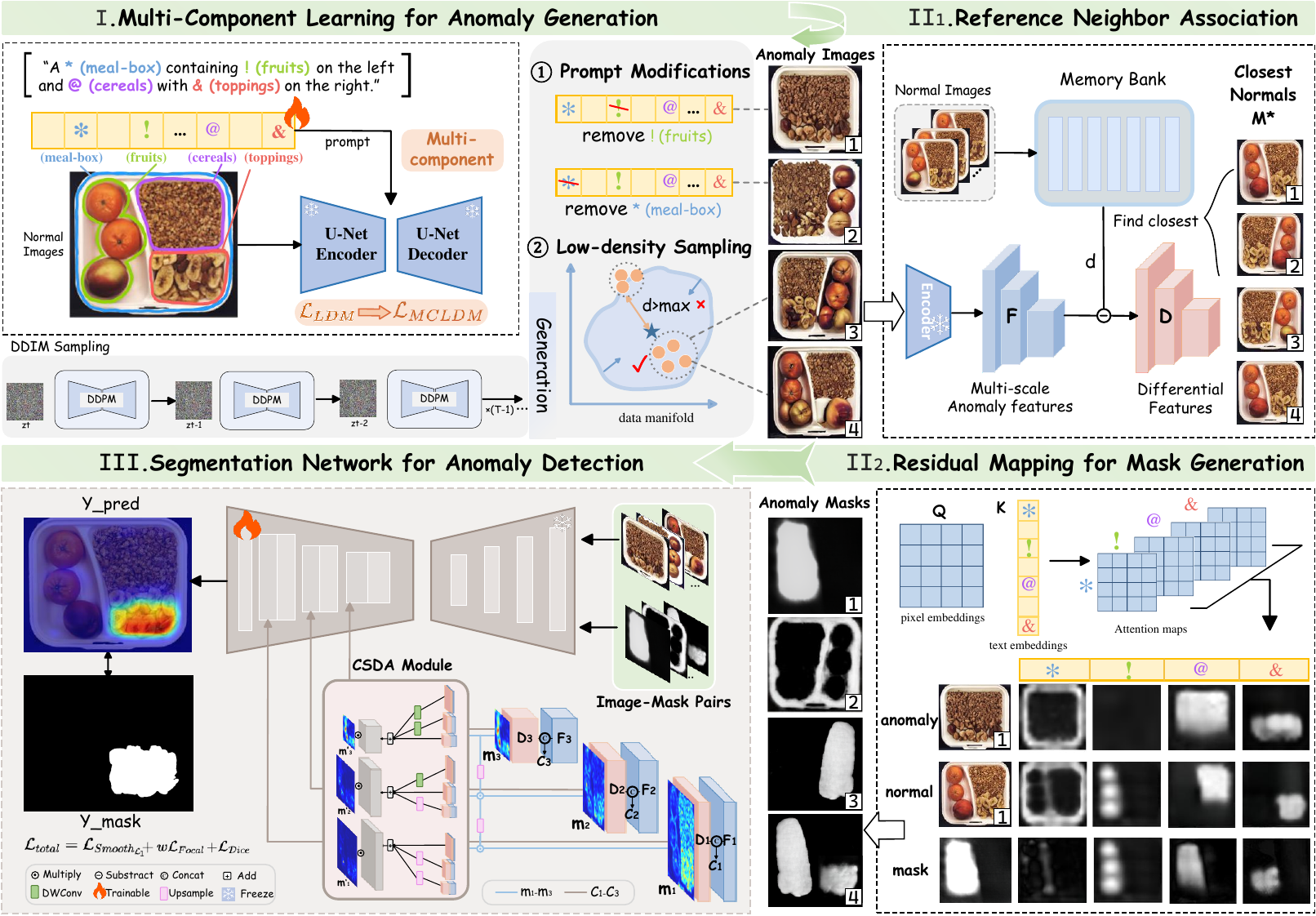} 
\caption{The pipeline of ComGEN consisting of three stages: Anomaly Generation, Mask Generation and Anomaly Detection. \textbf{I.} Multi-Component Learning (MCL) disentangles image regions to align text tokens and components. Then Prompt Modifications (PM) and Low-density Sampling (LS) enhance generation. \textbf{II.} Reference Neighbor Association (RNA) searches the closest normal samples to anomalies, which are input to Residual Mapping (RM) together to generated masks. \textbf{III.} Their differential features are fed to Cross-Scale Difference Aggregation Module (CSDA) for model acceleration.}
\label{fig:pipeline}
\end{figure*}

\section{METHOD}

Our method consists of three stages, as shown in Fig. \ref{fig:pipeline}. In the first stage, multiple text embeddings are learned from normal images, and high-quality anomalies are generated with two generation editing approaches. In the second stage, comparing with reference module, the generated anomalies iteratively search for closest normal neighbors to create anomaly masks based on their residual attention for decoupled components. In the third stage, the pair-wise anomalies and masks serve as supervision signals for a modified segmentation network to complete the anomaly detection task.

\subsection{Preliminaries}
\textbf{Stable Diffusion.} StableDiffusion (SD) consists of a variational autoencoder (VAE) \cite{kingma2013auto}, U-Net \cite{ronneberger2015u}, and text encoder. The VAE encoder $\varepsilon$ compresses the image $x$ to a latent $z$, which is perturbed by Gaussian noise $\epsilon\sim \text{Uniform}\left( 1,T \right)$ in the forward diffusion process. The U-Net, parameterized by $\theta$, denoises the noisy latent by predicting the noise. This denoising process can be conditioned on text prompts $y$ encoded by text encoder $\tau_{\theta}$. The training process is to minimize the loss function below: 
\begin{equation}
L_{\text{LDM}}=\mathbb{E}_{\varepsilon \left( x \right) ,\epsilon,y,t}\left[ ||\epsilon -\epsilon _{\theta}\left( \varepsilon \left( x \right) ,t,\tau _{\theta}\left( y \right) \right) ||_{2}^{2} \right] 
\end{equation}

\textbf{Text condition.} Let $y$ have $n$ tokens, 
$\tau _{\theta}\left( y \right)$ is mapped to intermediate feature maps through cross-attention layers as follows, where $A,Q,K,V$ denote attention, query, key and value matrices. $A\left[ i,j,k \right]$ represents the information flow from the $k$-th text token to the $(i,j)$ latent pixel. 
\begin{equation}
\text{attn}\left( Q,K,V \right) =A\cdot V=\text{soft}\max \left( \frac{QK^{\top}}{\sqrt{d}} \right) \cdot V
\end{equation}

\textbf{Textual Inversion.} Specially, given a single embedding $v$ learned by textual inversion \cite{gal2022image} from user-specified images $I_{a}$, which is then injected into diffusion model as fresh knowledge to achieve new-concept generation. 
\begin{equation}
v^*\!=\!\text{arg}\min_v\mathbb{E}_{\varepsilon \left( x \right) ,\epsilon ,y,t}\left[ ||\epsilon\! -\!\epsilon _{\theta}\!\left( \varepsilon \left( I_{a}^{t} \right) ,t,\tau _{\theta}\left( y \right) \right)\! ||_{2}^{2} \right] 
\end{equation}

\subsection{Anomaly Generation} \label{sec:ag}
\textbf{Multi-component Learning Strategy.} Since it is challenging for a single embedding in Textual Inversion \cite{gal2022image} to capture semantics of every component within a new-concept industrial image, we design a Multi-component learning strategy to learn multiple components simultaneously from a sentence-image pair. ComGEN learns a list of embeddings $\mathcal{V}\!=\!\left[ v^*,v^!,...,v^@,v^{\&} \right]$ corresponds to new components $\mathcal{C}\!=\!\left[ c^*,c^!,...,c^@,c^{\&} \right]$, initialized by $\tau _{\theta}$. Learnable tokens are arranged by adjectives, conjunctions and prepositions. The generation is conditioned on phrases constructed from templates derived from CLIP \cite{radford2021learning} to introduce diversity. 

A naive adaptation of \cite{gal2022image} to achieve multi-word learning is learning embeddings for all prompts in the string, but prompts like conjunctions and prepositions are irrelevant. To maintain a focused learning objective, we only set nouns as learnable embeddings to reduce network parameters. The optimization is still guided by the image-level LDMs, but now updating while keeping $\tau _{\theta}$ and $\epsilon _{\theta}$ frozen. Multi-component learning loss can be denoted as follows: 
\begin{equation}
L_{\text{MCLDM}}\!=\!\mathbb{E}_{\varepsilon \left( x \right) ,\epsilon ,y,t,\mathcal{V}}\left[ ||\epsilon -\epsilon _{\theta}\left( \varepsilon \left( I_{a}^{t} \right) ,t,c_{\mathcal{V}}\!\left( y \right) \right) \!||_{2}^{2} \right]
\end{equation}
where we simplify the expression $\left[ \tau _{\theta}\left( y \right) ,v^*,...,v^{\&} \right]$ as $c_{\mathcal{V}}\!\left( y \right)$, and $v^i \!= \!c_{v^i}\!\left( y \right)$ represents each learnable embedding vector. We insert each 768-parameter learnable embeddings into U-Net in LDMs \cite{rombach2022high} implementing with cross-attention.

\textbf{Manipulate Cross-attention Maps via Prompt Modifications.} Since our model learns the semantics of each component through learnable embeddings, it effectively disentangles the overall image into distinct components. As shown in Fig. \ref{fig:pipeline}II, cross-attention maps establish connections between each component region within the image and the corresponding tokens \cite{hertz2022prompt}, so we can manipulate cross-attention maps to simulate real-world logical anomalies. For example, 'missing' semantics often occur in the industry, so we can remove related tokens to replicate this scenario.

\textbf{Low-Density Anomalies Sampling under Distance Constraints.} Strong prior knowledge introduces diversity to the generated images by creating combinations of components that differ from the normal data. Due to anomalous images being located in low-density regions near the normal images, these generated samples can be considered as anomaly patterns that may appear in low-probability density regions of the data manifold. Fig. \ref{fig:pipeline}I shows that these samples simulate various anomaly patterns, such as changes in quantity, structural alterations, and color variations. However, the degree of their deviations from the normal data should not be too large ($d<{d_{max}}$)), to ensure that the paired masks can be generated accurately. The distance $d$ is defined as the distance between the anomalous image and its closest normal image in the feature space, which will be explained in Sec. \ref{sec:mg}.

\subsection{Mask Generation} \label{sec:mg}

\textbf{Reference Neighbor Association.} In the second stage, we generate anomaly masks that are paired with the anomaly images $\mathcal{Y}_A$ generated in the first stage. Comparable to the human learning process and embedding-based methods, the abnormal regions are obtained by comparing the anomaly image with the normal images in our memory. Therefore, we begin with finding the closest normal patterns of generated anomaly images. We first propose a reference module extended on memory bank to incorporate given $N$ reference normal images $\mathcal{M}{:}=\Big\{( x_{n},- ) \Big\}_{n=1}^{N}$. We store reference features at different scales, using $\mathcal{M}_j\in\mathbb{R}^{N\times c^j\times h^j\times w^j}(j\in\{1,2,3\})$ to denote the $j$-the block output of memory module from a pre-trained ResNet-like encoder such as ResNet-18 \cite{he2016deep,defard2021padim}, where $c^j,h^j,w^j$ represents depth, height and width of $j$-th scale feature maps. As shown in Fig. \ref{fig:pipeline}II, given a generated anomaly image $y_i$, the encoder also extracts multi-scale anomaly features to obtain $\mathcal{F}_{i,j}=F_{j}(y_{i}) (y_{i}\in \mathcal{Y}_{A})$. Then we can find the closest reference pattern $\mathcal{M}_{j}^{i,*}$ at $j$-th scale by calculating the L2 distance between $\mathcal{F}_{i,j}$ and each of the reference features $\mathcal{M}_{j}$. We define the closest reference sample $\mathcal{M}^{i,*}$ to the anomaly image $y_i$ as:
\begin{equation}
\begin{aligned}
\mathcal{D}_{i,j}=& D(\mathcal{F}_{i,j}-\mathcal{M}^{i,*}), \\
\mathrm{s.t.~}\mathcal{M}^{i,*}=\underset{\mathcal{M}^{n}\subset{\mathcal{M}}}{\operatorname*{argmin}}&\sum_{\mathcal{M}_{j}^{n}\subset{\mathcal{M}^n}}\|\mathcal{F}_{i,j}-\mathcal{M}_{j}^{n}\|_{2}  
\end{aligned}
\end{equation}
\begin{equation}
d_{i} = \sum\nolimits_{j=1}^{3} SSIM(\mathcal{F}_{i,j}, \mathcal{M}_{j}^{i,*})
\end{equation}
where $D(\cdot,\cdot)$ implements element-wise Euclidean distance, $\mathcal{D}_{i,j}$ are t he differential features between $\mathcal{F}_{i,j}$ and $\mathcal{M}^{i,*}$. $d_i$ represents the distance in Sec. \ref{sec:ag} and SSIM is computed channel-wise. $\mathcal{D}_{i,j}$ and $\mathcal{F}_{i,j}$ are fed into Fig. \ref{fig:pipeline}III.

\begin{figure*}[t]
\centering
\includegraphics[width=1\textwidth]{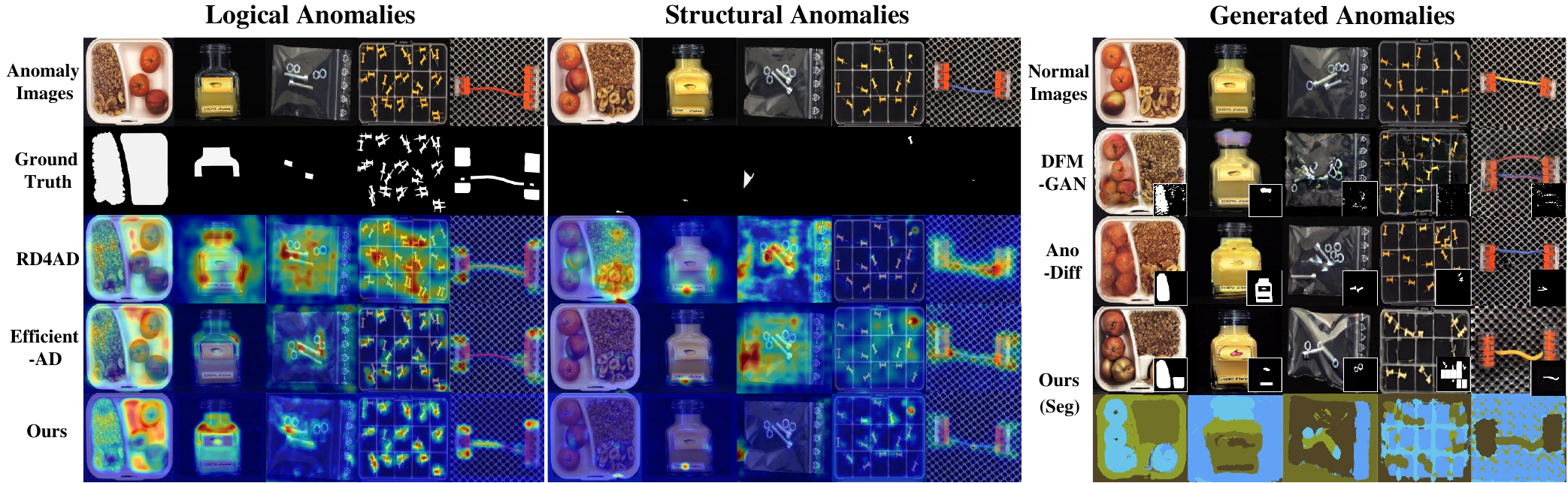} 
\caption{Localization and generation results of ComGEN from MVTecLOCO dataset. \textbf{Logical and Structural Anomalies:} Comparison of detection results between others and ours. \textbf{Generated Anomalies:} Images with masks (bottom-right). \textit{Seg} shows segmented components based on cross-attention maps.}
\label{fig:dandg}
\end{figure*}

\textbf{Text-to-Component Residual Mapping.} To tell the difference between generated anomaly images with their closest reference samples, given the input prompt $\mathcal{P}$, we consider $p$ learnable tokens $S=[*,!,\ldots,@,\&]$ present in $\mathcal{P}$ representing new components $\mathcal{C}\!=\!\left[ c^*,c^!,...,c^@,c^{\&} \right]$. Our objective is to extract a spatial attention map for each token $\mathrm{s\in S}$, indicating the influence of $s$ on each image patch. Given the noised latent $z_t$, we perform a forward pass through the pre-trained U-Net network \cite{ronneberger2015u} using $z_t$ and $\mathcal{P}$. Among the stored intermediate attention maps, we average all 16×16 attention layers and heads, which are proven to be the most semantically meaningful ones \cite{hertz2022prompt}. The resulting aggregated attention map $A_t$ contains $p$ spatial attention maps. As shown in Fig. \ref{fig:pipeline}, we select smoothed attention maps $A_0$ at time step $t=0$ as pixel-level annotations of $p$ components.
\begin{equation}
A_0=\left\{\mathrm{Gaussian}\left(A_0^s\right)\mid s=1,2,\ldots,p\right\}
\end{equation}
Here, $s$ denotes the index of the learnable tokens, and $\mathrm{Gaussian}(\cdot)$ utilizes a Gaussian kernel with a size of 3.

First, for generated anomaly images, $z_t$ is sampled from a standard normal distribution $\mathcal{N}(0,I)$, where $t$ denotes the current time step in the reverse diffusion process. The resulting cross-attention maps are denoted as $A_0^{gen}$.

Second, for the corresponding closest reference samples, to ensure that the generated images accurately reconstruct the reference images, we adopt DDIM process \cite{song2020denoising}. Specifically, we start with $\mathcal{M}_{0}^{*}=\mathcal{M}^{*}$ and the deterministic forward DDIM process to obtain latents is represented as:
\begin{equation}
\mathcal{M}_{t+1}^*=\sqrt{\alpha_{t+1}}f_\theta(\mathcal{M}_t^*,t)+\sqrt{1-\alpha_{t+1}}\epsilon_\theta(\mathcal{M}_t^*,t)
\label{eq:forward}
\end{equation}
and the deterministic reverse DDIM process to generate samples from the obtained latent becomes:
\begin{equation}
\mathcal{M}_{t\boldsymbol{-}1}^*=\sqrt{\alpha_{t\boldsymbol{-}1}}f_\theta(\mathcal{M}_t^*,t)+\sqrt{1-\alpha_{t\boldsymbol{-}1}}\epsilon_\theta(\mathcal{M}_t^*,t)
\label{eq:reverse}
\end{equation}
By applying Formula (\ref{eq:forward}) for $t\!\in\!\{0,...,T-1\}$, we can encode the closest normal sample $\mathcal{M}_0^*$ in a noisy image $\mathcal{M}_T^*$. Then, we recover the identical $\mathcal{M}_0^*$ from $\mathcal{M}_T^*$ by using Formula (\ref{eq:reverse}) for $t\in\{T,...,1\}$. During this process, cross-attention maps of the closest reference sample are recorded as $A_0^{norm}$.

Finally, we can calculate the residuals of the cross-attention maps between generated anomalies and closest reference samples to obtain anomaly masks $y^{mask}$. The $\mathrm{Upsample}(\cdot)$ adjusts the resolution to match the input image. 
\begin{equation}
y^{mask}=Upsample(\sum_{s=1}^{p}|A_{0}^{s,norm}-A_{0}^{s,gen}|)
\end{equation}

\subsection{Anomaly Detection}

\textbf{Cross-Scale Difference Aggregation Module.} We obtained concatenated information $\mathcal{C}_{i,j}$ composed of the input image features $\mathcal{F}_{i,j}$ and the best difference information $\mathcal{D}_{i,j}$ in Fig. \ref{fig:pipeline}II. We found that $\mathcal{D}_{i,j}$ already contains a blurred estimate of the abnormal region in the channel-wise averaged feature maps $m_{ij}$. By generating progressively enhanced masks $m_{ij}'$ from $m_{ij}$ through Formula (\ref{eq:mask}), the noise in the implicit anomalous estimate $m_{ij}$ is reduced in Fig. \ref{fig:pipeline}III. 

To facilitate the exchange of cross-scale information, inspired by \cite{wang2020deep}, multi-scale features $\mathcal{C}_{ij}$ are merged in U-Net structure. For downscaling, depth-wise separable convolutions are applied, and for upscaling, bilinear interpolation is followed by a 1x1 convolution. The segmentation network based on U-Net is in Fig. \ref{fig:pipeline}III, where $\mathcal{C}_{ij}$ masked by $m_{ij}$ are fed to the final decoder. The decoder outputs an anomaly score map $y^{pred}$, which is of the same shape as $y^{mask}$. 
\begin{equation}
m_{i3}'\!=m_{i3}\quad  
m_{i2}'\!=m_{i2}\odot m_{i3}'^u\quad
m_{i1}'\!=m_{i1}\odot m_{i2}'^u
\label{eq:mask}
\end{equation}
where $\odot$ element-wise product and \textit{u} represents up-sampling.

\textbf{Training Constraints.} Our training objective comprises three parts: Focal Loss \cite{lin2017focal} is applied to emphasize the hard misclassified examples; Smooth L1 Loss \cite{girshick2015fast} is to reduce the oversensitivity to outliers; while Dice Loss \cite{milletari2016v} is to address unbalanced segmentation training.
\begin{multline}
\mathcal{L}_{total}=Smooth_{{\mathcal{L}_1}}(y^{pred},y^{mask})+\omega\mathcal{L}_{Focal}(y^{pred},y^{mask}) \\
+\mathcal{L}_{Dice}(y^{pred},y^{mask})
\end{multline}

\section{EXPERIMENTS}

\subsection{Experimental Settings}

\textbf{Datasets.} We mainly conduct experiments on the MVTecLOCO dataset \cite{bergmann2022beyond}, a mainstream dataset for anomaly detection, including both structural and logical anomalies, with a total of 3,644 images distributed across five categories. 

\textbf{Evaluation Metrics.} We classify metrics into generation metrics and performance metrics. Generation metrics are dedicated to evaluating the quality of the generated images. Inception Score (IS) \cite{barratt2018note} is applied to access the quality of generated images while Intra-cluster Pairwise Learned Perceptual Image Patch Similarity (IC-LPIPS) metric \cite{ojha2021few} is to measure the diversity among the generated anomalies. On the other hand, performance metrics focus on assessing the effectiveness of the generated images in downstream tasks like anomaly detection and localization. We employ metrics such as the Area Under the Receiver Operating Characteristic curve (AUROC) and Saturated Per-Region Overlap (sPRO) \cite{zhao2024logical}, to evaluate the precision and overall efficacy of the proposed framework in these specific tasks. 

\textbf{Implementation Details.} We retain the original hyper-parameter choices in \cite{gal2022image} to train the first stage. We train our models for each category using one-third of the normal images over 6100 epochs on a V100 GPU, with a constant learning rate of 1e-5 and a batch size of 16. We adopt DDIM sampling method with the denoising step T = 50 to generate 1,000 images per category. Meanwhile, we perform random cropping and rotation on the image to get augmented views. $\omega$ in the total loss is set to 5. To construct the reference memory module, we get multi-scale features of dimensions of 64×64×64, 128×32×32, and 256×16×16 from block 1, block 2, and block 3 of ResNet-18 \cite{he2016deep}. 

\begin{table}[t]
\caption{Comparison of others and Ours on IS and IC-L indicators. Left/right represent logical/structural anomalies.}
\centering
\resizebox{\columnwidth}{!}{ 
\renewcommand{\arraystretch}{1.0} 
\small
\begin{tabular}{lcccccccc}
\toprule
\multirow{2}{*}{Category} & \multicolumn{2}{c}{DFMGAN\cite{duan2023few}} & \multicolumn{2}{c}{AnoDiff\cite{hu2024anomalydiffusion}} & \multicolumn{2}{c}{Ours} \\
\cmidrule(lr){2-3} \cmidrule(lr){4-5} \cmidrule(lr){6-7}
& IS↑ & IC-L↑ & IS↑ & IC-L↑ & IS↑ & IC-L↑ \\
\midrule
breakfast\_box        & 1.21/1.33 & 0.24/0.23 & 1.28/1.49 & 0.29/0.27 & \textbf{1.41/1.58} & \textbf{0.35/0.32} \\
juice\_bottle         & 1.12/1.17 & 0.13/0.13 & 1.21/1.25 & 0.18/0.16 & \textbf{1.35/1.45} & \textbf{0.22/0.20} \\
pushpins             & 1.02/1.14 & 0.23/0.24 & 1.08/1.31 & 0.37/\textbf{0.31} & \textbf{1.20/1.48} & \textbf{0.45/}0.30 \\
screw\_bag            & 1.34/1.49 & 0.20/0.21 & 1.67/1.73 & 0.26/0.25 & \textbf{1.98/2.14} & \textbf{0.32/0.30} \\
connectors  & 1.18/1.72 & 0.33/0.23 & 1.35/1.98 & 0.40/\textbf{0.30} & \textbf{1.82/2.24} & \textbf{0.45}/0.29 \\
\midrule
Average    & 1.17/1.37 & 0.23/0.21 & 1.32/1.55 & 0.30/0.26 & \textbf{1.55/1.78} & \textbf{0.36/0.28} \\
\bottomrule
\end{tabular}}
\label{tab:gen}
\end{table}

\begin{table*}[t]
\caption{Detection and Localization Results. Top: Image-level AU-ROC / Pixel-level AU-sPRO (FPR 5\%) on MVTecLOCO\cite{bergmann2022beyond}. \\ Bottom: AUC-I and AUC-P denote image-level and pixel-level AU-ROC for Diesel Engine and MVTecAD\cite{bergmann2019mvtec}.}
\centering
\resizebox{\textwidth}{!}{
\begin{tabular}{lccccccccccc}
\toprule
\multirow{2}{*}{Category} & \multicolumn{4}{c}{Methods W/O anomaly generation} & \multicolumn{6}{c}{Methods With anomaly generation} \\
\cmidrule(lr){2-5} \cmidrule(lr){6-11}
 & RD4AD\cite{deng2022anomaly} & GCAD\cite{bergmann2022beyond} & THFR\cite{guo2023template} & EfficientAD\cite{batzner2024efficientad} & DRAEM\cite{zavrtanik2021draem} & DFMGAN\cite{duan2023few} & AnoDiff\cite{hu2024anomalydiffusion} & GRAD\cite{dai2024generating} & LogicalAL\cite{zhao2024logical} & Ours\\
\midrule
breakfast\_box  & 68.7 / 42.2     & 83.9 / 50.2   & 78.0 / 58.3   & - / -      & 75.7 / 49.9   & 70.5 / 57.9   & 79.7 / \textbf{60.6}   & 81.2 / -    & 85.4 / 46.8     & \textbf{87.2} / 50.8 \\
juice\_bottle   & 94.8 / 85.1     & 99.4 / 91.0   & 97.1 / 89.6   & - / -      & 93.9 / 80.0   & 94.0 / 81.2   & 94.9 / 82.9   & 97.6 / -    & 98.5 / 91.3     & \textbf{99.1} / \textbf{93.3} \\
Pushpins      & 75.9 / 61.4     & 86.2 / 73.9   & 88.3 / 76.3   & - / -      & 76.0 / 49.3   & 78.5 / 58.8   & 84.4 / 65.8   & \textbf{99.7} / -    & 87.4 / \textbf{81.3}     & 92.9 / 79.3 \\
screw\_bag      & 74.9 / 57.4     & 63.2 / 55.8   & 73.7 / 61.5   & - / -      & 72.7 / 49.0   & 73.0 / 54.4   & 73.2 / \textbf{55.8}   & 76.6 / -    & \textbf{82.0} / 52.3     & 79.8 / 55.3 \\
connector    & 84.4 / 71.3     & 89.3 / 79.8   & 92.7 / 84.8   & - / -      & 82.5 / 67.3   & 85.9 / 70.1   & 86.7 / 74.8   & 85.4 / -    & 89.0 / 76.3     & \textbf{95.2} / \textbf{80.3} \\
\midrule
Mean          & 79.7 / 63.5     & 83.3 / 70.1   & 86.0 / 74.1   & 90.7 / 79.8   & 80.1 / 59.1   & 80.3 / 64.5   & 83.8 / 68.0   & 87.5 / -    & 88.5 / 69.7     & \textbf{91.2} / \textbf{71.8} \\
\bottomrule
\end{tabular}}

\vspace{0.2cm}

\resizebox{\textwidth}{!}{
\renewcommand{\arraystretch}{0.9}
\begin{tabular}{cccccccccccccc}
\toprule
\multirow{2}{*}{Method} & \multicolumn{5}{c}{Diesel Engine (AUC-I / AUC-P on Four Resolutions)} & \multirow{2}{*}{Method} & \multicolumn{6}{c}{MVTec AD (AUC-P on Five Categories)} \\
\cmidrule(lr){2-6} \cmidrule(lr){8-13}
 & 150×150 & 200×200 & 300×300 & 600×600 & Average &  & Carpet & Grid & Leather & Tile & Wood & Average\\
\midrule
DRAEM \cite{zavrtanik2021draem}    & 54.4 / 77.4     & 57.1 / 80.2   & 57.1 / 82.2   & 47.0 / 87.1  & 53.9 / 81.7  & MCDEN \cite{yang2023multi}   & 97.5   & 96.6   & \textbf{99.3}   & 97.8    & 91.7     & 96.6 \\
Ours     & \textbf{64.3 / 85.9}     & \textbf{65.8 / 88.7}   & \textbf{61.4 / 89.0}   & \textbf{63.9 / 87.7}  & \textbf{63.9 / 87.8}  & Ours    & \textbf{98.3}   & \textbf{97.2}   & 98.8   & \textbf{97.9}    & \textbf{98.5}     & \textbf{98.1} \\
\bottomrule
\end{tabular}}
\label{tab:loc}


\end{table*}

\subsection{Comparative Experiments and Main Results}

\textbf{Anomaly generation results.} We mainly compare our model with a represented GAN-based method DFMGAN \cite{duan2023few} and latest Diffusion-based method AnoDiff \cite{hu2024anomalydiffusion}. Tab. \ref{tab:gen}. demonstrates that our model achieves the highest scores in both IS and IC-LPIPS, indicating that it generates anomalous data with superior quality and diversity. Furthermore, we exhibit the generated anomalies in Fig. \ref{fig:dandg}. DFMGAN occasionally produces identical masks and anomalies due to its inherent instability. AnoDiff struggles to generate defects that align with the mask because it lacks an understanding of logical anomalies when integrating defects into normal images. In contrast, our method is capable of generating realistic and diverse defects beyond the scope of the test set, while also producing accurately paired masks. Our approach provides more comprehensive and diverse defect data for downstream detection tasks in an unsupervised way.

\textbf{Anomaly detection and classification results.} Tab. \ref{tab:loc} presents the results for image-level AU-ROC and pixel-level AU-sPRO, highlighting that our model surpasses other anomaly generation models. Visualized localization results are shown in Fig. \ref{fig:dandg}. For classification, our approach achieved an average classification accuracy of 52.2\%, significantly surpassing DFMGAN's 36.1\% and AnoDiff's 45.4\% on both logical and structural anomalies.

\subsection{Ablation Study}
We investigate the importance of each module in ComGEN and the results are reported in Tab. \ref{tab:ablation}. We employ each setting to generate 1000 anomalous image-mask pairs to report performance metrics. 
1) MCL: we compare the effects of single-prompt learning and the multi-component learning strategies. Simply applying a textual inversion \cite{gal2022image} model tends to capture overall objects, conversely, learning multiple prompts enables model to disentangle multiple components within an image; 2) PM and LS: significantly boosted realism and diversity compared to randomly sampling; 3) RNA: as the number of memory samples increases, the model locates the abnormal regions more accurately, but excessive memory samples leads to decreased inference speed. 4) CSDA: removing CSDA causes the degradation of performance, indicating that it is necessary to introduce implicit multi-scale anomalous information to accelerate model learning.

\begin{table}[htbp]
\caption{Ablation study on each module in ComGEN. IS/IC-L are influenced by MCL and PM/LS Module.}
\centering
\label{tab:ablation}
\resizebox{\columnwidth}{!}{
\renewcommand{\arraystretch}{0.90} 
\begin{tabular}{ccccccc}
\toprule
\textbf{MCL} & \textbf{PM/LS} & \textbf{RNA} & \textbf{CSDA} & \textbf{AU-ROC/AU-sPRO} & \textbf{IS / IC-L} \\
\midrule
             &                &             &             & 81.8 / 64.7            & 1.25 / 0.22 \\
\checkmark   &                &             &             & 87.2 / 67.1            & 1.54 / 0.29 \\
             &                & \checkmark  & \checkmark  & 83.6 / 65.0            & - / - \\
\checkmark   & \checkmark     &             &             & 89.2 / 69.5            & - / - \\
\checkmark   &                & \checkmark  & \checkmark  & 88.5 / 68.9            & - / - \\
\checkmark   & \checkmark     & \checkmark  &             & 89.9 / 70.4            & - / - \\
\checkmark   & \checkmark     & \checkmark  & \checkmark  & \textbf{91.2 / 71.8}   & \textbf{1.67 / 0.32} \\
\bottomrule
\end{tabular}}
\label{tab:ablation}
\renewcommand{\arraystretch}{1.0} 
\end{table}

\subsection{Real-World Applications on Diesel Engine Industry }
\begin{figure}[t]
\centering
\includegraphics[width=1\columnwidth]{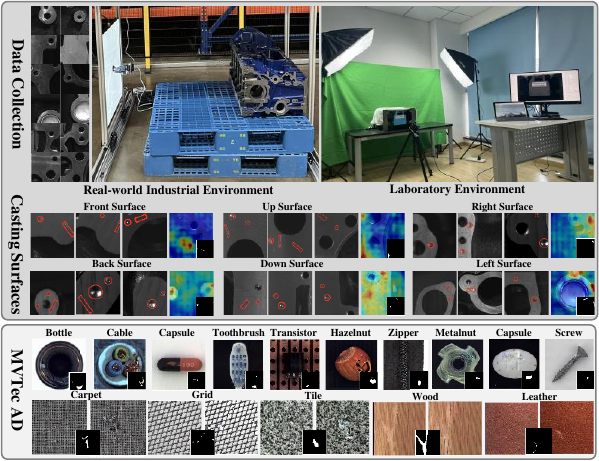} 
\caption{Generation and localization results. \textbf{Data Collection:} Hardware-based Data Acquisition Environment. \textbf{Casting Surfaces:} Multi-surface results on Diesel Engine, where generated anomalies are marked with red boxes and heap maps highlight anomalous regions (ground-truth masks in low-right corner). \textbf{MVTec AD:} Generation results of object categories (from Bottle to Screw) and texture categories (from Carpet to Leather), where we compared texture anomalies with Anodiff (left) and Ours (right).}
\label{fig4}
\end{figure}

We conducted experiments on Diesel Engine both in real-world industry scenario and laboratory environment to validate the effectiveness of anomaly generation. Fig. \ref{fig4} depicts the environment in which we collected data on casting surfaces in a Diesel Engine Manufacturing Factory. We apply the settings in \cite{hu2024anomalydiffusion} and only executed the first stage of training and generation. Following the DRAEM \cite{zavrtanik2021draem} anomaly detection process, we replaced 30\% of the generation data with anomalous images generated by ComGEN. This resulted in a \textbf{10.0 / 6.1} increase in AUROC metric in Tab. \ref{tab:loc}. As shown in Fig.4, our method is especially effective to generate mixture anomalies. Furthermore, we also experiment on MVTecAD \cite{bergmann2019mvtec}, a widely-used anomaly detection dataset. Our method increases the AUROC from 97.3 to 97.9 for all categories and achieves \textbf{98.1} on five texture categories, compared to MCDEN's 96.6 \cite{yang2023multi} in Tab. \ref{tab:loc}. The results in Fig. \ref{fig4} further underscore the potential of our method to enhance detection performance for generating realistic anomalies in the industrial automation manufacturing.

\section{CONCLUSIONS}
We introduce a Component-aware Generation (ComGEN) algorithm for anomaly detection, with only normal samples to generate anomalous data. In summary, Multi-Component Learning strategy effectively disentangles visual components in text, facilitating the generation of realistic anomalies through Prompt Modifications and Low-Density Sampling. The integration of Reference Neighbor Association and text-to-component Residual Mapping, produces residual attention and multi-scale differential features to generate mask and detect anomalies with Cross-Scale Difference Aggregation Module. ComGEN hypothesizes that logical anomalies can be disentangled by aligning each token's attention with its corresponding component in the object regions. Experiments on MVTecLOCO dataset and applications on real-world Diesel Engine demonstrate the effectiveness of our model, to generate authentic anomalies for industrial manufacturing. 

\section*{ACKNOWLEDGMENT}
This work was supported by National Natural Science Foundation of China (No.62406075), National Key Research and Development Program of China (2023YFC3604802), National Natural Science Foundation of China (No.62072112).






\bibliographystyle{IEEEtran}  

\begin{thebibliography}{10}
\providecommand{\url}[1]{#1}
\csname url@rmstyle\endcsname
\providecommand{\newblock}{\relax}
\providecommand{\bibinfo}[2]{#2}
\providecommand\BIBentrySTDinterwordspacing{\spaceskip=0pt\relax}
\providecommand\BIBentryALTinterwordstretchfactor{4}
\providecommand\BIBentryALTinterwordspacing{\spaceskip=\fontdimen2\font plus
\BIBentryALTinterwordstretchfactor\fontdimen3\font minus \fontdimen4\font\relax}
\providecommand\BIBforeignlanguage[2]{{%
\expandafter\ifx\csname l@#1\endcsname\relax
\typeout{** WARNING: IEEEtran.bst: No hyphenation pattern has been}%
\typeout{** loaded for the language `#1'. Using the pattern for}%
\typeout{** the default language instead.}%
\else
\language=\csname l@#1\endcsname
\fi
#2}}

\bibitem{mei2018unsupervised}
S.~Mei, H.~Yang, and Z.~Yin, ``An unsupervised-learning-based approach for automated defect inspection on textured surfaces,'' \emph{IEEE transactions on instrumentation and measurement}, vol.~67, no.~6, pp. 1266--1277, 2018.

\bibitem{long2021fabric}
X.~Long, B.~Fang, Y.~Zhang, G.~Luo, and F.~Sun, ``Fabric defect detection using tactile information,'' in \emph{2021 IEEE International Conference on Robotics and Automation (ICRA)}.\hskip 1em plus 0.5em minus 0.4em\relax IEEE, 2021, pp. 11\,169--11\,174.

\bibitem{yu2024multi}
J.~Yu, C.~B. Chi, S.~Fichera, P.~Paoletti, D.~Mehta, and S.~Luo, ``Multi-class road defect detection and segmentation using spatial and channel-wise attention for autonomous road repairing,'' \emph{arXiv preprint arXiv:2402.04064}, 2024.

\bibitem{zavrtanik2021draem}
V.~Zavrtanik, M.~Kristan, and D.~Sko{\v{c}}aj, ``Draem-a discriminatively trained reconstruction embedding for surface anomaly detection,'' in \emph{Proceedings of the IEEE/CVF international conference on computer vision}, 2021, pp. 8330--8339.

\bibitem{zhang2023destseg}
X.~Zhang, S.~Li, X.~Li, P.~Huang, J.~Shan, and T.~Chen, ``Destseg: Segmentation guided denoising student-teacher for anomaly detection,'' in \emph{Proceedings of the IEEE/CVF Conference on Computer Vision and Pattern Recognition}, 2023, pp. 3914--3923.

\bibitem{deng2022anomaly}
H.~Deng and X.~Li, ``Anomaly detection via reverse distillation from one-class embedding,'' in \emph{Proceedings of the IEEE/CVF conference on computer vision and pattern recognition}, 2022, pp. 9737--9746.

\bibitem{roth2022towards}
K.~Roth, L.~Pemula, J.~Zepeda, B.~Sch{\"o}lkopf, T.~Brox, and P.~Gehler, ``Towards total recall in industrial anomaly detection,'' in \emph{Proceedings of the IEEE/CVF conference on computer vision and pattern recognition}, 2022, pp. 14\,318--14\,328.

\bibitem{zhang2021defect}
G.~Zhang, K.~Cui, T.-Y. Hung, and S.~Lu, ``Defect-gan: High-fidelity defect synthesis for automated defect inspection,'' in \emph{Proceedings of the IEEE/CVF Winter Conference on Applications of Computer Vision}, 2021, pp. 2524--2534.

\bibitem{duan2023few}
Y.~Duan, Y.~Hong, L.~Niu, and L.~Zhang, ``Few-shot defect image generation via defect-aware feature manipulation,'' in \emph{Proceedings of the AAAI Conference on Artificial Intelligence}, vol.~37, no.~1, 2023, pp. 571--578.

\bibitem{hu2024anomalydiffusion}
T.~Hu, J.~Zhang, R.~Yi, Y.~Du, X.~Chen, L.~Liu, Y.~Wang, and C.~Wang, ``Anomalydiffusion: Few-shot anomaly image generation with diffusion model,'' in \emph{Proceedings of the AAAI Conference on Artificial Intelligence}, vol.~38, no.~8, 2024, pp. 8526--8534.

\bibitem{jiang2024cagen}
B.~Jiang, Y.~Xie, J.~Li, N.~Li, Y.~Jiang, and S.-T. Xia, ``Cagen: Controllable anomaly generator using diffusion model,'' in \emph{ICASSP 2024-2024 IEEE International Conference on Acoustics, Speech and Signal Processing (ICASSP)}.\hskip 1em plus 0.5em minus 0.4em\relax IEEE, 2024, pp. 3110--3114.

\bibitem{li2021cutpaste}
C.-L. Li, K.~Sohn, J.~Yoon, and T.~Pfister, ``Cutpaste: Self-supervised learning for anomaly detection and localization,'' in \emph{Proceedings of the IEEE/CVF conference on computer vision and pattern recognition}, 2021, pp. 9664--9674.

\bibitem{rombach2022high}
R.~Rombach, A.~Blattmann, D.~Lorenz, P.~Esser, and B.~Ommer, ``High-resolution image synthesis with latent diffusion models,'' in \emph{Proceedings of the IEEE/CVF conference on computer vision and pattern recognition}, 2022, pp. 10\,684--10\,695.

\bibitem{lee2024text}
M.~Lee and J.~Choi, ``Text-guided variational image generation for industrial anomaly detection and segmentation,'' in \emph{Proceedings of the IEEE/CVF Conference on Computer Vision and Pattern Recognition}, 2024, pp. 26\,519--26\,528.

\bibitem{bergmann2022beyond}
P.~Bergmann, K.~Batzner, M.~Fauser, D.~Sattlegger, and C.~Steger, ``Beyond dents and scratches: Logical constraints in unsupervised anomaly detection and localization,'' \emph{International Journal of Computer Vision}, vol. 130, no.~4, pp. 947--969, 2022.

\bibitem{bergmann2019mvtec}
P.~Bergmann, M.~Fauser, D.~Sattlegger, and C.~Steger, ``Mvtec ad--a comprehensive real-world dataset for unsupervised anomaly detection,'' in \emph{Proceedings of the IEEE/CVF conference on computer vision and pattern recognition}, 2019, pp. 9592--9600.

\bibitem{zavrtanik2021reconstruction}
V.~Zavrtanik, M.~Kristan, and D.~Sko{\v{c}}aj, ``Reconstruction by inpainting for visual anomaly detection,'' \emph{Pattern Recognition}, vol. 112, p. 107706, 2021.

\bibitem{liu2023simplenet}
Z.~Liu, Y.~Zhou, Y.~Xu, and Z.~Wang, ``Simplenet: A simple network for image anomaly detection and localization,'' in \emph{Proceedings of the IEEE/CVF Conference on Computer Vision and Pattern Recognition}, 2023, pp. 20\,402--20\,411.

\bibitem{deng2009imagenet}
J.~Deng, W.~Dong, R.~Socher, L.-J. Li, K.~Li, and L.~Fei-Fei, ``Imagenet: A large-scale hierarchical image database,'' in \emph{2009 IEEE conference on computer vision and pattern recognition}.\hskip 1em plus 0.5em minus 0.4em\relax Ieee, 2009, pp. 248--255.

\bibitem{niu2020defect}
S.~Niu, B.~Li, X.~Wang, and H.~Lin, ``Defect image sample generation with gan for improving defect recognition,'' \emph{IEEE Transactions on Automation Science and Engineering}, vol.~17, no.~3, pp. 1611--1622, 2020.

\bibitem{schluter2022natural}
H.~M. Schl{\"u}ter, J.~Tan, B.~Hou, and B.~Kainz, ``Natural synthetic anomalies for self-supervised anomaly detection and localization,'' in \emph{European Conference on Computer Vision}.\hskip 1em plus 0.5em minus 0.4em\relax Springer, 2022, pp. 474--489.

\bibitem{dtd2014accuracy}
L.~Sharan, R.~Rosenholtz, and E.~H. Adelson, ``Accuracy and speed of material categorization in real-world images,'' \emph{Journal of vision}, vol.~14, no.~9, pp. 12--12, 2014.

\bibitem{goodfellow2020generative}
I.~Goodfellow, J.~Pouget-Abadie, M.~Mirza, B.~Xu, D.~Warde-Farley, S.~Ozair, A.~Courville, and Y.~Bengio, ``Generative adversarial networks,'' \emph{Communications of the ACM}, vol.~63, no.~11, pp. 139--144, 2020.

\bibitem{zhang2023adding}
L.~Zhang, A.~Rao, and M.~Agrawala, ``Adding conditional control to text-to-image diffusion models,'' in \emph{Proceedings of the IEEE/CVF International Conference on Computer Vision}, 2023, pp. 3836--3847.

\bibitem{zhao2024logical}
Y.~Zhao, ``Logical: Towards logical anomaly synthesis for unsupervised anomaly localization,'' in \emph{Proceedings of the IEEE/CVF Conference on Computer Vision and Pattern Recognition}, 2024, pp. 4022--4031.

\bibitem{dai2024generating}
S.~Dai, Y.~Wu, X.~Li, and X.~Xue, ``Generating and reweighting dense contrastive patterns for unsupervised anomaly detection,'' in \emph{Proceedings of the AAAI Conference on Artificial Intelligence}, vol.~38, no.~2, 2024, pp. 1454--1462.

\bibitem{kingma2013auto}
D.~P. Kingma, ``Auto-encoding variational bayes,'' \emph{arXiv preprint arXiv:1312.6114}, 2013.

\bibitem{ronneberger2015u}
O.~Ronneberger, P.~Fischer, and T.~Brox, ``U-net: Convolutional networks for biomedical image segmentation,'' in \emph{Medical image computing and computer-assisted intervention--MICCAI 2015: 18th international conference, Munich, Germany, October 5-9, 2015, proceedings, part III 18}.\hskip 1em plus 0.5em minus 0.4em\relax Springer, 2015, pp. 234--241.

\bibitem{gal2022image}
R.~Gal, Y.~Alaluf, Y.~Atzmon, O.~Patashnik, A.~H. Bermano, G.~Chechik, and D.~Cohen-Or, ``An image is worth one word: Personalizing text-to-image generation using textual inversion,'' \emph{arXiv preprint arXiv:2208.01618}, 2022.

\bibitem{radford2021learning}
A.~Radford, J.~W. Kim, C.~Hallacy, A.~Ramesh, G.~Goh, S.~Agarwal, G.~Sastry, A.~Askell, P.~Mishkin, J.~Clark, \emph{et~al.}, ``Learning transferable visual models from natural language supervision,'' in \emph{International conference on machine learning}.\hskip 1em plus 0.5em minus 0.4em\relax PMLR, 2021, pp. 8748--8763.

\bibitem{hertz2022prompt}
A.~Hertz, R.~Mokady, J.~Tenenbaum, K.~Aberman, Y.~Pritch, and D.~Cohen-Or, ``Prompt-to-prompt image editing with cross attention control,'' \emph{arXiv preprint arXiv:2208.01626}, 2022.

\bibitem{he2016deep}
K.~He, X.~Zhang, S.~Ren, and J.~Sun, ``Deep residual learning for image recognition,'' in \emph{Proceedings of the IEEE conference on computer vision and pattern recognition}, 2016, pp. 770--778.

\bibitem{defard2021padim}
T.~Defard, A.~Setkov, A.~Loesch, and R.~Audigier, ``Padim: a patch distribution modeling framework for anomaly detection and localization,'' in \emph{International Conference on Pattern Recognition}.\hskip 1em plus 0.5em minus 0.4em\relax Springer, 2021, pp. 475--489.

\bibitem{song2020denoising}
J.~Song, C.~Meng, and S.~Ermon, ``Denoising diffusion implicit models,'' \emph{arXiv preprint arXiv:2010.02502}, 2020.

\bibitem{wang2020deep}
J.~Wang, K.~Sun, T.~Cheng, B.~Jiang, C.~Deng, Y.~Zhao, D.~Liu, Y.~Mu, M.~Tan, X.~Wang, \emph{et~al.}, ``Deep high-resolution representation learning for visual recognition,'' \emph{IEEE transactions on pattern analysis and machine intelligence}, vol.~43, no.~10, pp. 3349--3364, 2020.

\bibitem{lin2017focal}
T.-Y. Lin, P.~Goyal, R.~Girshick, K.~He, and P.~Doll{\'a}r, ``Focal loss for dense object detection,'' in \emph{Proceedings of the IEEE international conference on computer vision}, 2017, pp. 2980--2988.

\bibitem{girshick2015fast}
R.~Girshick, ``Fast r-cnn,'' in \emph{Proceedings of the IEEE international conference on computer vision}, 2015, pp. 1440--1448.

\bibitem{milletari2016v}
F.~Milletari, N.~Navab, and S.-A. Ahmadi, ``V-net: Fully convolutional neural networks for volumetric medical image segmentation,'' in \emph{2016 fourth international conference on 3D vision (3DV)}.\hskip 1em plus 0.5em minus 0.4em\relax Ieee, 2016, pp. 565--571.

\bibitem{barratt2018note}
S.~Barratt and R.~Sharma, ``A note on the inception score,'' \emph{arXiv preprint arXiv:1801.01973}, 2018.

\bibitem{ojha2021few}
U.~Ojha, Y.~Li, J.~Lu, A.~A. Efros, Y.~J. Lee, E.~Shechtman, and R.~Zhang, ``Few-shot image generation via cross-domain correspondence,'' in \emph{Proceedings of the IEEE/CVF conference on computer vision and pattern recognition}, 2021, pp. 10\,743--10\,752.

\bibitem{guo2023template}
H.~Guo, L.~Ren, J.~Fu, Y.~Wang, Z.~Zhang, C.~Lan, H.~Wang, and X.~Hou, ``Template-guided hierarchical feature restoration for anomaly detection,'' in \emph{Proceedings of the IEEE/CVF International Conference on Computer Vision}, 2023, pp. 6447--6458.

\bibitem{batzner2024efficientad}
K.~Batzner, L.~Heckler, and R.~K{\"o}nig, ``Efficientad: Accurate visual anomaly detection at millisecond-level latencies,'' in \emph{Proceedings of the IEEE/CVF Winter Conference on Applications of Computer Vision}, 2024, pp. 128--138.

\bibitem{yang2023multi}
H.~Yang, H.~Zhu, J.~Li, J.~Chen, and Z.~Yin, ``Multi-category decomposition editing network for the accurate visual inspection of texture defects,'' \emph{IEEE Transactions on Automation Science and Engineering}, 2023.

\end{thebibliography}

\end{document}